%% file: acl_latex.tex
\documentclass[11pt]{article}

\usepackage[final]{acl}
\usepackage{times}
\usepackage{latexsym}
\usepackage{booktabs}
\usepackage{amsfonts}
\usepackage{boxedminipage}
\usepackage[T1]{fontenc}
\usepackage{amsmath}
\usepackage[utf8]{inputenc}

\usepackage{microtype}

\usepackage{inconsolata}

\usepackage{graphicx}
\usepackage{xcolor,colortbl}

\usepackage{amssymb}
\usepackage{amsthm}
\usepackage{tabularx}
\usepackage{fancyvrb}

\usepackage{subcaption}
\usepackage{multirow}

\usepackage{algorithm,algpseudocode}
\usepackage{booktabs}

\title{Relational Over-Regularization: Graph-Based AI-Generated Text Detection via Sentence Transition Deviation}

\author{Hyeonchu Park, Bugeun Kim
\\
        Department of Artificial Intelligence, Chung-Ang University, Republic of Korea \\
        \{phchu0429, bgnkim\}@cau.ac.kr}

\begin{document}
\maketitle
\begin{abstract}
Detecting AI-generated text (AIGT) remains challenging because existing approaches rely on token-level statistical signals or independent stylometric features, causing them to overfit to specific generators and fail under distribution shift. We identify a structural signal at the sentence-pair level: LLMs produce inter-sentence transition variance that deviates from human writing through \textit{inflated variance} driven by recurring similarity bursts at paragraph boundaries and templated transitions. We formalize this as \textit{Relational Over-Regularization} (ROR) and validate it across four benchmarks ($p < 0.001$). The central contribution is this relational problem formulation, not a novel GNN architecture; CSFG is one concrete instantiation for operationalizing ROR. To exploit this signal, we propose the \textbf{C}ross-Source \textbf{S}tylometric \textbf{F}ingerprint \textbf{G}raph (CSFG), a graph-based framework that encodes positional, sequential, semantic, and transition deviation signals as learnable GNN edge features. The per-edge signed deviation $\delta_{ij}$ operationalizes ROR without hand-crafted thresholds and acts as a false-positive calibrator. CSFG achieves 97.14\% accuracy under binary detection, outperforming the strongest graph-based baseline by 11.14 pp, with a false positive rate of 1.57\% and robust generalization to unseen LLMs in the inflated-variance regime; detection degrades for generators whose transition variance falls at or below the human baseline.
\end{abstract}

\input{section/introduction}
\input{section/background}
\input{section/method}
\input{section/experiment}
\input{section/result}
\input{section/conclusion}
\input{section/limitation}

\section*{Acknowledgments}
This research was supported by Basic Science Research Program through the National Research Foundation of Korea(NRF) funded by the Ministry of Education (RS-2025-25434151) and the Institute of Information \& Communications Technology Planning \& Evaluation (IITP) grant funded by the Korea government (MSIT) [RS-2021-II211341, Artificial Intelligence Graduate School Program (Chung-Ang University)].


\bibliography{custom}

\input{section/appendix}

\end{document}

%% file: section/introduction.tex
\section{Introduction}
Recent advances in Large Language Models (LLMs) have enabled AI-generated text (AIGT) to reach a level of fluency indistinguishable from human writing, raising concerns about misinformation, ghostwriting, and academic misconduct. Existing detection approaches, perplexity-based statistical signals \citep{detectgpt,hansen} and Transformer-based classifiers \citep{hc3,f-detectgpt}, share two limitations: they overfit to generator-specific surface distributions and degrade under unseen conditions LLMs \citep{raid}, and they treat linguistic features as independent variables, failing to capture sentence-level relational structure.

We argue that the key structural signal lies at the sentence-pair level. Human writing exhibits organically irregular inter-sentence similarity distributions from topic drift, affective variation, and revision inconsistency. LLMs optimized via next-token prediction and RLHF \citep{feed}, by contrast, predominantly produce \textbf{recurring similarity bursts}, sharp local spikes at paragraph boundaries, topic restatements, and templated transitions, that inflate and pattern-skew transition variance in a manner systematically distinguishable from human writing. While \citet{self} observed document-level semantic stability in AIGT, we identify a complementary pattern: structured local bursts define the dominant relational signature of AI-generated discourse across the generators we study. We formalize this as \textbf{Relational Over-Regularization (ROR)} and validate it empirically across four benchmarks (\S\ref{sec:ror}). Importantly, ROR constitutes the central problem formulation of this work: we frame AIGT detection as identifying deviations in relational transition structure, rather than introducing another architecture-specific detection mechanism. CSFG implements this formulation, designed to operationalize ROR through learnable edge representations. We note, however, that certain generators deviate from this pattern in the opposite direction—producing \textit{hyper-uniform} text whose transition variance falls at or below the human baseline, a boundary condition whose implications for the detector scope we discuss in the context of our generalization experiments.

To exploit this signal, we represent each document as a \textbf{sentence graph} whose edges carry a \textbf{transition deviation} feature $\delta_{ij}$, the signed difference between a sentence pair's cosine similarity and the document-level mean. A positive $\delta_{ij}$ flags a local similarity burst; a negative value indicates a suppressed transition.
This per-edge feature captures the inflated-variance regime of ROR without document-level aggregation or hand-crafted thresholds, while the signed encoding preserves sensitivity to both directional deviations for generators observed during training. We propose \textbf{Cross-Source Stylometric Fingerprint Graph (CSFG)}, a GNN that detects AIGT from the structural pattern of transition deviations across all edges.

Our contributions are: (1)~\textbf{ROR}, a sentence-pair-level structural signal of AIGT validated across four benchmarks; (2)~\textbf{transition deviation} $\delta_{ij}$, a novel per-edge GNN feature encoding local similarity burst signatures; (3)~\textbf{CSFG}, a graph-based framework jointly modeling sequential and semantic sentence relations with $\delta_{ij}$ as a learnable edge feature; and (4)~97.14\% binary detection accuracy, outperforming the strongest graph-based baseline by 11.14pp with low false positive rates on unseen LLMs in the inflated-variance regime.

%% file: section/background.tex
\section{Related Work}
\label{sec:related}

\subsection{AIGT Detection}

AIGT detection (AIGTD) has emerged as a major research problem with the rapid advancement of LLMs~\cite{survey}. Existing approaches fall into two categories: perplexity-based zero-shot detection~\cite{detectgpt, f-detectgpt, hansen} and supervised Transformer classifiers~\cite{gpt-roberta, radar}. Despite strong benchmark performance, the RAID~\citep{raid} and
M4~\citep{m4} benchmarks show that detection degrades substantially under unseen LLMs, domain shifts, and adversarial rewriting, suggesting over-reliance on token-level surface signals rather than discourse-level regularity patterns.

Recent work has explored structural consistency as a detection signal.
\citet{self} demonstrated that AIGT exhibits more stable semantic structures than human text via masking-based self-consistency analysis. However, macro-level stability cannot capture finer-grained sentence-pair phenomena: Recurring local bursts at paragraph boundaries may be absorbed into a globally stable mean, remaining invisible to aggregate analysis. We identify this sentence-pair transition signal as a complementary discriminator and formalize it in~\S\ref{sec:method}.

\subsection{Discourse Structure and Relational Stylometry}
\label{sec:related_discourse}

Traditional stylometry summarises text as an independent document-level scalar features~\citep{survey2}, which, while discriminative~\citep{turing}, are structurally blind to \emph{relational} signals: whether a given transition is anomalous relative to the document's own distributional baseline. Discourse coherence modelling~\citep{modeling, probabilistic} establishes that textual naturalness emerges from relational structures between sentences, motivating sentence-pair transitions as the primary detection unit.

We bridge these traditions using the transition deviation $\delta_{ij}$: the signed difference between a sentence pair's cosine similarity and the document-level mean, capturing the extent to which each transition departs from the document's baseline. Human text exhibits organic irregularity in this deviation, whereas LLMs produce structurally biased patterns (Relational Over-Regularization;~\S\ref{sec:ror})—a signal captured by neither independent stylometric features nor absolute coherence scores.

\subsection{Graph-based Text Representation Learning}
\label{sec:related_graph}

GNNs have been widely adopted in NLP for relational text modelling~\citep{gcn,survey3}. For AIGT detection, CoCo~\citep{coco}
combines coherence graphs with contrastive learning at the sentence level, but encodes coherence as an absolute scalar similarity without capturing the signed deviation $\delta_{ij}$ from the document mean—and thus cannot distinguish local similarity bursts from uniformly high similarity, precisely the distinction that characterizes AI-generated discourse under ROR. \citet{gnn_detection} exploits syntactic dependency graphs to improve detection performance, but focuses on token-level syntactic structures rather than sentence-level semantic transition patterns. More recently, \citet{motif} proposes LM$^2$otifs, an explainable GNN framework that constructs word co-occurrence graphs and extracts interpretable motifs to
differentiate human and machine-generated text; however, operating at the word level, it does not model inter-sentence transition dynamics and thus cannot capture the document-level relational variance that characterizes ROR.
We address this gap with a sentence-level graph that encodes transition deviation as an explicit edge feature.

%% file: section/method.tex
\section{Method}
\label{sec:method}

\begin{figure*}[t]
\centering
  \includegraphics[width=1.8\columnwidth]{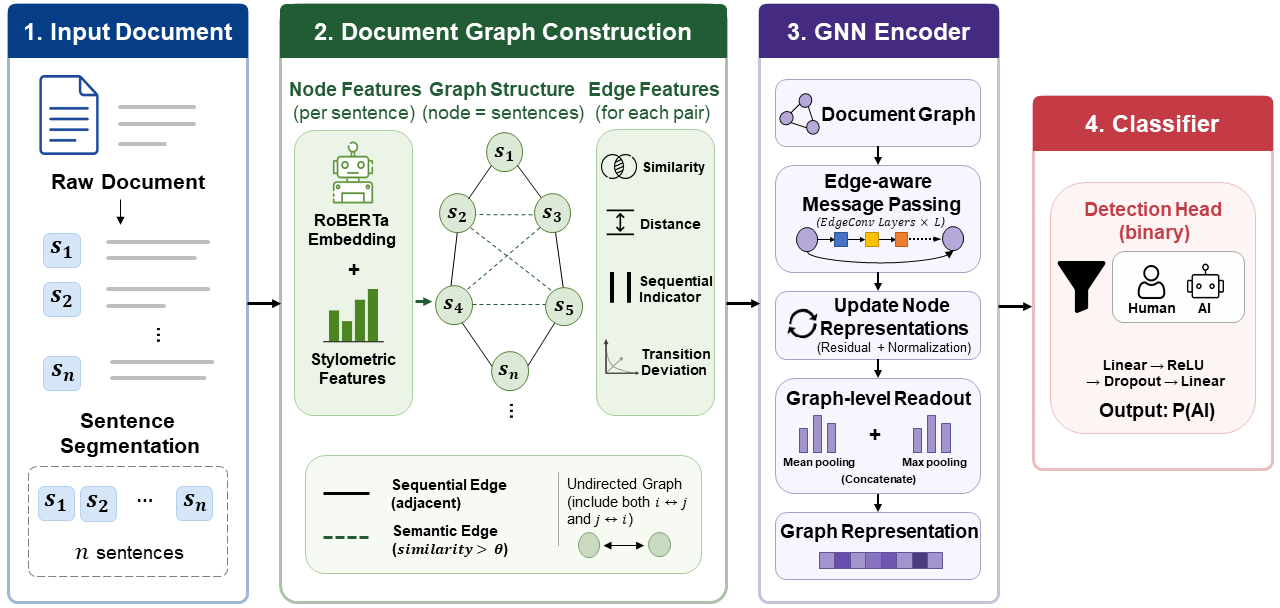}
  \caption{Overview of the CSFG framework.}
  \label{fig:method}
\end{figure*}

We propose the \textbf{Cross-Source Stylometric Fingerprint Graph}
(\textbf{CSFG}), a GNN-based framework that models sentence-level relational dynamics for AIGTD. The \emph{cross-source} property reflects joint training on human-written text and multiple LLM outputs, with an auxiliary source-attribution head that discourages generator-specific surface artifacts. A GNN is chosen because the distinguishing signature of AIGT lies not in individual sentence properties but in \emph{inter-sentence transition patterns}—the structurally biased variance of sentence-to-sentence similarity. By representing sentences as nodes and pairwise relationships as typed, feature-rich edges, the GNN aggregates local and long-range relational context through iterative message passing, an inductive bias unavailable to
independent sentence classifiers or sequence models.

\subsection{Hypothesis: Relational Over-Regularization}
\label{sec:ror}

LLMs optimized for next-token prediction and RLHF produce discourse with a \textbf{structurally biased transition variance pattern}~\citep{feed}.
This deviation manifests in two directions:
(1)~\textbf{inflated variance} driven by \textbf{recurring similarity bursts}, sharp local spikes from paragraph boundaries, topic restatements, and templated transitions; or (2)~\textbf{suppressed variance} (hyper-uniform generation), where inter-sentence similarity is held below the organic irregularity of human discourse.
Human writing, by contrast, exhibits organic irregularity from topic drift, affective fluctuation, and revision inconsistency.

We attribute this structural bias to two complementary training pressures.
Next-token prediction encourages the model to maintain local semantics coherence by generating tokens that are strongly consistent with the preceding context, producing high-similarity plateaus within topically coherent spans.
RLHF further reinforces structured discourse organization: human raters reward responses with clear paragraph boundaries, explicit topic restatements, and summary sentences, which are perceived as well-organized and easy to follow~\citep{feed}.
These two pressures operate jointly: next-token prediction creates intra-span similarity plateaus, while RLHF reward shaping introduces sharp cross-span spikes at structurally salient boundaries.
The result is a transition variance pattern that is both higher in mean and more burst-skewed than the organic irregularity of human discourse, a signal we formalize as \textbf{Relational Over-Regularization (ROR)}.

Formally, let $\mathcal{T}(D) = \{\mathrm{sim}(s_i, s_{i+1})\}_{i=1}^{n-1}$ denote consecutive-sentence cosine similarities for document $D$.
The distribution of $\mathrm{Var}(\mathcal{T}(D))$ differs systematically between AIGT and human text in a direction determined by the generator's training objective and decoding procedure.
Critically, this signal is \emph{invisible to document-level mean similarity} $\bar{\sigma}(D)$: two documents may share an identical $\bar{\sigma}$ while differing markedly in local burst structure, as recurring spikes are absorbed into the global average~\citep{self}.
$\mathrm{Var}(\mathcal{T}(D))$ exposes this structure directly, making it a complementary discriminator to macro-level stability analysis.
As $\delta_{ij}$ is a \emph{signed} deviation from the document mean (Eq.~\ref{eq:delta}), it captures both regimes without explicit directional thresholds; however, generalization to hyper-uniform generators depends on their presence during training (see Appendix~\ref{app:generator_tv}).
We empirically validate the inflated-variance direction across four benchmarks in~\S\ref{sec:ror_empirical}.

\subsection{Document Graph Construction}
\label{sec:graph}

Each document $D$ is a graph $\mathcal{G} = (\mathcal{V}, \mathcal{E})$ with sentences as nodes and edges encoding relational dependencies.

\paragraph{Node Features.}
Each node $v_i$ concatenates a frozen \texttt{roberta-base} \texttt{[CLS]} embedding~\citep{rob} with a 10-dimensional stylometric vector:
\begin{equation}
  \mathbf{h}_i^{(0)} = \bigl[
    \mathbf{h}_i^{\mathrm{RoBERTa}}
    \;\|\;
    \mathbf{h}_i^{\mathrm{style}}
  \bigr] \in \mathbb{R}^{778}.
  \label{eq:node}
\end{equation}
The stylometric vector covers five discourse-function features (hedge-word, modal-verb, first-person pronoun ratios; question/exclamation marks) and five surface-complexity features (token count, type-token ratio, comma density, discourse-marker presence, long-word ratio), all of which have been shown to differentiate human and LLM-generated text~\citep{style,bb,cc}.

\paragraph{Sequential Edges.}
Adjacent sentences are connected to model a linear discourse trajectory:
$e_{i,i+1}^{\mathrm{seq}} \in \mathcal{E},\ i = 1,\ldots,n-1$.

\paragraph{Semantic Edges.}
Long-range dependencies are captured by connecting sentence pairs whose cosine similarity exceeds the threshold $\theta$:
\begin{equation}
  e_{ij}^{\mathrm{sem}} \in \mathcal{E}
  \quad \text{if} \quad
  \mathrm{sim}(s_i, s_j) > \theta,\ i < j.
  \label{eq:sem_edge}
\end{equation}
We set $\theta = 0.6$ by validation AUROC; sensitivity analysis is in Appendix~\ref{app:ab2}.

\paragraph{Edge Features.}
Each edge carries a four-dimensional vector:
\begin{equation}
  \mathbf{e}_{ij} = \bigl[\,
    \mathrm{sim}(s_i,s_j),\;
    \tilde{d}_{ij},\;
    \mathbb{1}_{\mathrm{seq}},\;
    \delta_{ij}
  \,\bigr] \in \mathbb{R}^{4},
  \label{eq:edge_feat}
\end{equation}
where $\tilde{d}_{ij}=|i-j|/(n-1)$ is normalised positional distance,
$\mathbb{1}_{\mathrm{seq}}=\mathbf{1}[|i-j|=1]$ flags sequential edges, and
\begin{equation}
  \delta_{ij} = \mathrm{sim}(s_i,s_j) - \bar{\sigma}, \quad
  \bar{\sigma} = \tfrac{1}{\binom{n}{2}}\sum_{i<j}\mathrm{sim}(s_i,s_j).
  \label{eq:delta}
\end{equation}
Computing $\bar{\sigma}$ over all pairs (rather than consecutive pairs only) captures the global similarity baseline against which each local burst is measured. Feature contributions are validated in Appendix~\ref{app:ab3}.

\subsection{GNN Encoder and Graph-level Readout}
\label{sec:gnn}

Node features are projected to a $d_h$-dimensional space, then updated over $L=3$ EdgeConv layers~\citep{edgeconv}:
\begin{align}
  \mathbf{h}_i^{\mathrm{proj}} &=
    \mathrm{ReLU}\!\left(\mathbf{W}_{\mathrm{in}}\mathbf{h}_i^{(0)}\right),
    \label{eq:proj}\\
  \mathbf{m}_{ij}^{(\ell)} &= \mathrm{ReLU}\!\left(
    \mathbf{W}_{\mathrm{msg}}^{(\ell)}
    \bigl[\mathbf{h}_i^{(\ell)}\|\mathbf{h}_j^{(\ell)}\|\mathbf{e}_{ij}\bigr]
  \right), \label{eq:msg}\\
  \mathbf{h}_i^{(\ell+1)} &= \mathrm{BN}\!\left(
    \mathbf{W}_{\mathrm{upd}}^{(\ell)}\Bigl[
      \mathbf{h}_i^{(\ell)}\Big\|
      \tfrac{1}{|\mathcal{N}(i)|}\textstyle\sum_{j\in\mathcal{N}(i)}
      \mathbf{m}_{ij}^{(\ell)}
    \Bigr]
  \right). \label{eq:update}
\end{align}
Because $\delta_{ij}$ enters every message (Eq.~\ref{eq:msg}), transition deviation modulates all layers without explicit thresholds. The graph representation concatenates mean and max pooling over final-layer nodes:
\begin{equation}
  \mathbf{h}_{\mathcal{G}} = \Bigl[
    \tfrac{1}{|\mathcal{V}|}\textstyle\sum_i\mathbf{h}_i^{(L)}
    \;\Big\|\;
    \max_i\mathbf{h}_i^{(L)}
  \Bigr] \in \mathbb{R}^{2d_h},
  \label{eq:readout}
\end{equation}
preserving complementary global (mean) and local (max) relational signals~\citep{power}. This is passed to a binary detection head $f_{\mathrm{bin}}:\mathbb{R}^{2d_h}\to\mathbb{R}^{2}$; an auxiliary source attribution head $f_{\mathrm{src}}:\mathbb{R}^{2d_h}\to\mathbb{R}^{K}$ is used only during training and discarded at inference. Both heads are
two-layer MLPs (linear–ReLU–dropout–linear).

\subsection{Training Objective}
\label{sec:training}

The model minimizes a joint objective:
\begin{equation}
  \mathcal{L} = \mathcal{L}_{\mathrm{bin}}
              + \lambda\,\mathcal{L}_{\mathrm{src}},
  \label{eq:loss}
\end{equation}
where $\mathcal{L}_{\mathrm{bin}}$ is binary cross-entropy and $\mathcal{L}_{\mathrm{src}}$ is a cross-entropy source attribution loss that acts as a domain-adversarial regulariser, preventing the encoder from collapsing to generator-specific surface cues. We set $\lambda=1.0$ (sensitivity analysis in Appendix~\ref{app:ab1}); full implementation details are in Appendix~\ref{sec:impl}.

%% file: section/experiment.tex
\section{Experiments}
\label{sec:exp}

We evaluate CSFG under three settings:
(1) \textbf{binary detection} (binary human vs.\ AI classification),
(2) \textbf{unseen model generalization} (trained on known LLMs, tested on
held-out generators), and
(3) \textbf{robustness evaluation} (paraphrasing, humanization, and
back-translation perturbations applied to AI-generated text, with no retraining).
Prior to the main detection experiments, we empirically validate the ROR
hypothesis by comparing $\mathrm{Var}(\mathcal{T}(D))$ distributions between
human-written and AI-generated text across all four benchmarks
(\S\ref{sec:ror_empirical}).
All experiments are run 10 times; we report the mean Accuracy (primary metric),
False Positive Rate (FPR) and False Negative Rate (FNR) across runs.
Accuracy is chosen as the primary metric because the evaluation sets are
balanced; FPR and FNR are reported to distinguish the social costs
of false accusations of human authors (FPR: human text incorrectly classified
as AI-generated) from missed detections (FNR: AI-generated text incorrectly
classified as human).
All supervised models use a 70/15/15 train/validation/test split;
The checkpoint with the highest validation AUROC is selected for the test
evaluation.

\paragraph{Datasets.}
We use four benchmarks:
\textbf{HC3}~\citep{hc3} (expert vs.\ ChatGPT, professional domains),
\textbf{AIGTBench}~\citep{aigtbench} (social-media style, 12 LLMs),
\textbf{M4}~\citep{m4} (multi-domain, multi-generator robustness), and
\textbf{MULTITuDE}~\citep{multitude} (8 LLMs).
Following \citet{motif}, we use up to 2{,}000 samples per generator for
Setting~1; Settings~2 and~3 use 1{,}000 samples in total.

\paragraph{Evaluation Settings.}
For Setting~1 (binary detection), we compare \textbf{CSFG} against five
baselines: the perplexity-based methods Fast-DetectGPT~\citep{f-detectgpt}, DNAGPT~\citep{dnagpt}, LASTDE~\citep{lastde}, Likelihood~\citep{gltr}
and , the supervised Transformer classifiers
RoBERTa~\citep{gpt-roberta}, RADAR~\citep{radar},DetectAnyLLM~\citep{any}, DeTeCtive~\citep{detective}, and SeqXGPT~\citep{seqx} 

and the graph-based CoCo~\citep{coco} with contrastive learning.
All supervised baselines are retrained from scratch under identical data
conditions to ensure a fair comparison.

For Setting~2 (unseen model generalization), we construct a separate test
set entirely excluded from training, using three recent LLMs:
\textbf{Claude Sonnet 4.6}~\citep{Claude46},
\textbf{GPT-5}~\citep{gpt5}, and
\textbf{Gemini 2.5 Flash Lite}~\citep{gemini25}.
Each generator contributes 500 AI-generated samples paired with a shared set
of 500 human-written documents, yielding balanced binary test sets per
generator.
All Setting~2 evaluations use the model trained in Setting~1 without any
retraining or fine-tuning on the held-out generator data.

For Setting~3 (robustness evaluation), we apply three perturbation conditions
to \textbf{Claude Sonnet 4.6}, \textbf{GPT-5} and \textbf{Gemini 2.5 Flash Lite} -generated text: paraphrasing, humanization,
and back-translation.
We use Claude Sonnet 4.6 as the perturbation engine for all three conditions.
We acknowledge that using the same model for generation and humanization is
a limitation, but our aim is to measure detector robustness under realistic conditions
LLM-based rewriting rather than stylistic humanization per se.
The detector is not retrained: the model from Setting~1 is applied directly
to perturbed texts, making this a zero-shot out-of-distribution evaluation.
Human-written texts are also perturbed to isolate whether performance changes
reflect detector robustness or a shift in the decision boundary; implications
are discussed in \S\ref{sec:robust}.

Text generation and perturbation follow a unified prompt protocol; full
prompts are provided in Appendix~\ref{app:prompt}.

%% file: section/result.tex
\section{Results and Discussion}

\subsection{Empirical Validation of Relational Over-Regularization}
\label{sec:ror_empirical}
\input{table/ab0}
\begin{figure*}[t]
  \centering
  \includegraphics[width=2\columnwidth]{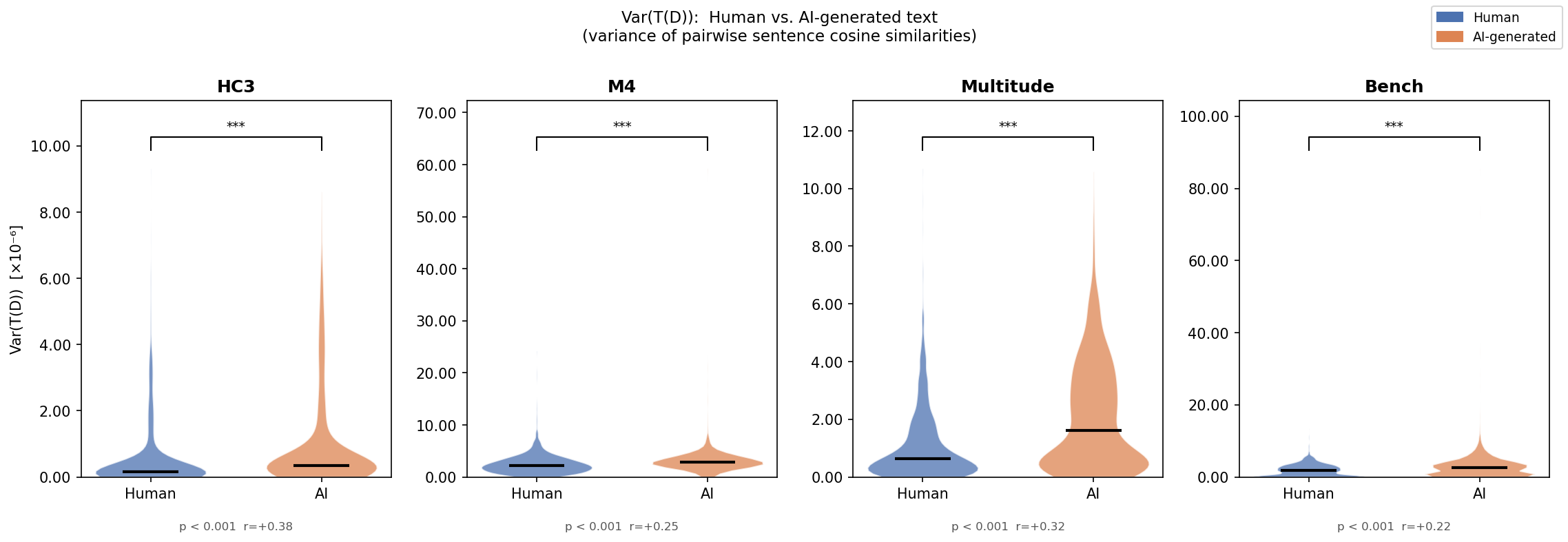}
  \caption{Distribution of $\mathrm{Var}(\mathcal{T}(D))$ for human-written (blue) and AI-generated (orange) text across four benchmarks, as violin plots. The horizontal bar: median. AIGT consistently exhibits higher variance and a heavier upper tail than human-written text, driven by recurring similarity
bursts at paragraph boundaries and templated transitions. Significance: $^{***}p < 0.001$. Effect sizes ($r$) and exact $p$-values
below each panel.}
  \label{fig:ror}
\end{figure*}

We empirically validate the ROR hypothesis by measuring $\mathrm{Var}(\mathcal{T}(D))$ for each document and comparing the resulting distributions between human-written and AIGT, using 1,000 samples per group across all four benchmarks. Statistical significance is assessed via the Mann-Whitney $U$ test; effect size is reported as rank-biserial correlation $r$. Results are summarised in Table~\ref{tab:ror} and Figure~\ref{fig:ror}.

AIGT consistently exhibits higher $\mathrm{Var}(\mathcal{T}(D))$ than human-written text across all four benchmarks ($p < 0.001$ in all cases), with effect sizes ranging from moderate on HC3 ($r = {+}0.38$) and MULTITuDE ($r = {+}0.32$) to small on M4 ($r = {+}0.25$) and AIGTBench ($r = {+}0.22$), confirming the inflated-variance regime of the ROR hypothesis for the generators represented in our training benchmarks. The distributional difference manifests not only in higher mean variance but also in a characteristically heavier upper tail (Figure~\ref{fig:ror}), reflecting recurring similarity bursts that are rare in human-written documents. As $\delta_{ij}$ encodes a \emph{signed} deviation from the document-level mean, the GNN captures
structural departure from human transition variance in both directions, including the hyper-uniform suppression observed in some frontier generators (Appendix~\ref{app:generator_tv}), without explicit directional thresholds.

\subsection{Binary Detection}
\label{sec:binary}

\input{table/result1-1}

Table~\ref{tab:main} reports binary detection performance across all baselines under Setting~1. CSFG achieves the highest accuracy of 97.14\% and the lowest FPR of 1.57\%, outperforming all baselines while maintaining a low FNR of 4.15\%. The low FPR indicates that $\delta_{ij}$ effectively suppresses false accusations of human-written text, consistent with its calibration role observed in the ablation study (\S\ref{app:ab3}).

Among the baselines, SeqXGPT is the strongest competitor, achieving 93.02\% accuracy and the lowest FNR of 1.53\%, but with a substantially higher FPR of 12.43\%. RoBERTa achieves the lowest baseline FPR of 5.85\% but exhibits the highest FNR of 28.55\%, indicating a conservative decision boundary. The graph-based baseline CoCo achieves 86.00\% accuracy with an FNR of 5.00\%, but its FPR remains high at 23.00\%, suggesting that discourse-level coherence alone is insufficient without explicitly modeling transition deviations.

Compared with SeqXGPT, CSFG improves accuracy by 4.12 percentage points while reducing the FPR from 12.43\% to 1.57\%. Compared with the graph-based baseline CoCo, CSFG further improves accuracy by 11.14 percentage points, reduces FPR by 21.43 percentage points, and lowers FNR by 0.85 percentage points. These improvements reflect the combined contributions of the four edge features—positional distance, sequential adjacency, cosine similarity, and transition deviation—together with the source-attribution regularizer. As confirmed by the ablation study (\S\ref{app:ab3}), positional distance is the primary contributor to detection accuracy, whereas $\delta_{ij}$ mainly improves calibration by reducing FPR without affecting overall accuracy. Finally, the consistently low standard deviation across 10 runs ($\sigma \leq 0.54$ pp for all metrics) demonstrates the stability of the proposed framework.

\subsection{Unseen Model Generalization}
\input{table/result2}

Table~\ref{tab:unseen} reports detection performance on three unseen LLMs: Claude Sonnet 4.6, GPT-5, and Gemini 2.5 Flash Lite. Performance varies substantially across generators, indicating that cross-model generalization remains highly model-dependent. CSFG achieves the highest accuracy on Claude Sonnet 4.6 and Gemini 2.5 Flash Lite, reaching 89.45\% and 92.26\%, respectively, while consistently maintaining the lowest FPR of 0.66\%. These results suggest that the proposed relational representation generalizes well to unseen generators exhibiting transition-variance patterns similar to those observed during training.

Among the baselines, LASTDE, Likelihood, and DetectAnyLLM perform close to random guessing, with accuracies generally below 65\% and unstable FPR/FNR trade-offs across generators. Recent learning-based methods, including DeTeCtive and SeqXGPT, exhibit more balanced behavior but remain below 62\% accuracy on all three unseen generators. CoCo is the strongest baseline, achieving competitive performance on Claude Sonnet 4.6 and Gemini 2.5 Flash Lite, although its FPR remains consistently at around 18\%, substantially higher than CSFG's.

GPT-5 presents a markedly different challenge. Both CSFG and CoCo achieve the same accuracy of 68.85\%, but their error profiles differ considerably. CoCo yields an FPR of 18.06\% with an FNR of 44.37\%, whereas CSFG reduces the FPR to only 0.66\% at the expense of an FNR of 61.64\%, indicating that it misses a large proportion of GPT-5-generated documents. As discussed in Appendix~\ref{app:generator_tv}, this limitation is not merely a calibration issue but stems from the proposed ROR hypothesis: GPT-5 exhibits a hyper-uniform transition-variance pattern that falls outside the variance regime learned during training. Additional experiments in Appendix~\ref{app:gpt5_train} further demonstrate that incorporating GPT-5 into training improves in-distribution detection but does not fully resolve this limitation.

Overall, the results demonstrate that CSFG consistently minimizes false accusations of human-written text while outperforming existing methods on generators that preserve the inflated transition-variance characteristic. At the same time, the GPT-5 results highlight an important limitation of the current framework and motivate future work on modeling both inflated-variance and hyper-uniform generation patterns.

\subsection{Robustness Against Text Perturbation}
\label{sec:robust}
\input{table/result3.tex}

Table~\ref{tab:robustness} reports detection performance under three perturbation settings: paraphrasing, humanization, and back-translation. Since the same overall trend is consistently observed across different generators, we present results for Claude Sonnet 4.6 in the main paper for brevity, while the corresponding results for GPT-5 and Gemini 2.5 Flash Lite are provided in Appendix~\ref{app:full}. No retraining is performed; the detector trained under Setting~1 is directly applied to perturbed texts, making this a zero-shot robustness evaluation.

Unlike Settings~1 and~2, the FPR in this setting is measured on perturbed human-written text. Therefore, it reflects the extent to which perturbation causes human documents to resemble AI-generated text and is not directly comparable to the FPR reported under standard evaluation.

Most baseline detectors exhibit substantial performance degradation under perturbation. LASTDE, Likelihood, and DetectAnyLLM achieve accuracies below 75\% across all settings, while DeTeCtive and SeqXGPT show improved robustness but remain sensitive to rewriting strategies, particularly under humanization. Among all baselines, CoCo consistently achieves the strongest performance, obtaining accuracies of 91.55\%, 89.22\%, and 89.55\% for paraphrasing, humanization, and back-translation, respectively.

CSFG remains competitive across all perturbation settings, achieving the highest accuracy under back-translation at 90.96\% and performance comparable to CoCo under paraphrasing (90.59\% vs.\ 91.55\%) and humanization (87.43\% vs.\ 89.22\%). Similar to CoCo, CSFG exhibits substantially elevated FPRs under paraphrasing and humanization while maintaining consistently low FNRs between 1.22\% and 5.10\%. As discussed in Appendix~\ref{sec:robust}, back-translation produces a substantially lower FPR than direct stylistic rewriting, suggesting that it better preserves the original transition structure.

We note a limitation of the humanization setting: Claude Sonnet 4.6 is used as both the text generator and the humanization model, which may preserve generator-specific relational patterns. A more rigorous evaluation would employ human rewriting or a model from a different family, which we leave for future work.

These robustness results assume ordinary-length documents drawn from conventional benchmarks. We additionally test two boundary conditions outside this regime: documents under 500 characters, and human-written text from a domain with formulaic discourse structure (news). Under both conditions, FPR rises sharply---to 64--71\% on short documents for two of three unseen generators, and to 25.00\% on structured human news text (XSum; \citealp{xsum}), against 0.66\% and 1.57\% respectively under standard evaluation. We report the full results and analysis in Appendix~\ref{app:full}, since
they identify conditions under which CSFG's low-FPR advantage does not hold.

%% file: table/ab0.tex
\begin{table*}[t]
\centering
\begin{tabular}{lcccccc}
\toprule
\textbf{Dataset} & $n_{\text{human}}$ & $n_{\text{AI}}$ 
  & $\mu_{\text{human}}$ & $\mu_{\text{AI}}$ 
  & $p$-value & $r$ \\
\midrule
HC3       &  1000 &  1000 
  & $5.72 \times 10^{-7}$ & $1.05 \times 10^{-6}$ 
  & $3.34 \times 10^{-45}$  & $+0.38$ \\
M4        &  1000 & 1000 
  & $2.52 \times 10^{-6}$ & $3.06 \times 10^{-6}$ 
  & $5.21 \times 10^{-22}$  & $+0.25$ \\
MULTITuDE & 1000 &  1000 
  & $1.13 \times 10^{-6}$ & $2.12 \times 10^{-6}$ 
  & $9.73 \times 10^{-60}$  & $+0.32$ \\
AIGTBench     & 1000 & 1000 
  & $1.95 \times 10^{-6}$ & $2.84 \times 10^{-6}$ 
  & $1.34 \times 10^{-73}$  & $+0.22$ \\
\bottomrule
\end{tabular}
\caption{Mann--Whitney $U$ test results for $\mathrm{Var}(\mathcal{T}(D))$
across four benchmarks. $\mu$ denotes the per-group mean variance.
$r$: rank-biserial correlation as effect size.
Positive $r$ indicates $\mathrm{Var}(\mathcal{T}(D))$ is higher in
AI-generated text than in human-written text.
All differences are statistically significant ($p < 0.001$).}
\label{tab:ror}
\end{table*}

%% file: table/result1-1.tex
\begin{table}[h]
\centering
\small
\begin{tabular}{lccc}
\toprule
\multirow{1}{*}{\textbf{Detector}} 
  & \multicolumn{1}{c}{\textbf{ACC (\%)}} 
  & \multicolumn{1}{c}{\textbf{FPR (\%)}} 
  & \multicolumn{1}{c}{\textbf{FNR (\%)}} \\

\midrule
F-DetectGPT  & 70.45  & 17.30  & 17.30 \\
DNAGPT          & 81.23  & 24.60  & 19.15  \\

LASTDE   &  58.93          & 38.82         &  43.31        \\
Likelhood  &   60.75        &   36.96     &   41.54     \\
DetectAnyLLM &  64.56          &   18.65       &  52.23        \\
DeTeCtive &  87.78          & 6.74         &   17.70       \\
SeqXGPT &   93.02        & 12.43          & 1.53        \\

RoBERTa        $^\dagger$ & 82.80   &  5.85  & 28.55   \\
RADAR           & 61.38  & 28.65  & 24.23  \\

CoCo            & 86.00  & 23.00  &  5.00  \\
\midrule
\textbf{CSFG (Ours)} & \textbf{97.14} & \textbf{1.57} & \textbf{4.15} \\
\bottomrule
\end{tabular}
\caption{Binary detection performance under standard evaluation (Setting 1). 
$^\dagger$RoBERTa is evaluated in an API-only setting via the Hugging Face inference endpoint.}
\label{tab:main}
\end{table}

%% file: table/result2.tex
\begin{table*}[h]
\centering

\begin{tabular}{llccc}
\toprule
\textbf{Detector} & \textbf{Generator} 
  & \textbf{Accuracy (\%)} 
  & \textbf{FPR (\%)} 
  & \textbf{FNR (\%)} \\
\midrule
\multirow{3}{*}{LASTDE}
  & Claude Sonnet 4.6     &50.50	&72.00	&26.40  \\
  & GPT-5                   &51.80	&89.00	&6.10     \\
  & Gemini 2.5 Flash Lite  &52.30	&18.50	&77.80     \\
\midrule

\multirow{3}{*}{Likelihood}
  & Claude Sonnet 4.6    &55.50	&68.20	&20.10   \\
  & GPT-5                 &55.70	&37.40	&51.40    \\
  & Gemini 2.5 Flash Lite   &52.10	&39.00	&57.10      \\
\midrule

\multirow{3}{*}{DetectAnyLLM}
  & Claude Sonnet 4.6   &64.00	&29.90	&42.10      \\
  & GPT-5                   &60.80	&18.70	&59.80     \\
  & Gemini 2.5 Flash Lite &54.00	&19.50	&72.60  \\
\midrule

\multirow{3}{*}{DeTeCtive}
  & Claude Sonnet 4.6   &52.30	&46.20	&49.40    \\
  & GPT-5                  &56.40	&48.40	&38.60     \\
  & Gemini 2.5 Flash Lite &56.50	&32.30	&55.10      \\
\midrule

\multirow{3}{*}{SeqXGPT}
  & Claude Sonnet 4.6    &61.20	&33.30	&44.50     \\
  & GPT-5                 &57.90	&42.70	&41.50      \\
  & Gemini 2.5 Flash Lite  &56.90	&32.50	&54.10     \\
\midrule

\multirow{3}{*}{CoCo}
  & Claude Sonnet 4.6     & 85.83 & 18.06   &  10.38 \\
  & GPT-5                  & 68.85 & 18.06  &  44.37 \\
  & Gemini 2.5 Flash Lite  & 89.74 & 17.89  & 3.19 \\
\midrule

\multirow{3}{*}{\textbf{CSFG (Ours)}}
  & Claude Sonnet 4.6       & \textbf{89.45} & 0.66 & 20.46 \\
  & GPT-5          & \textbf{68.85} & 0.66 & 61.64 \\
  & Gemini 2.5 Flash Lite  & \textbf{92.26} & 0.66 &  14.81 \\
\bottomrule
\end{tabular}
\caption{Detection performance on unseen LLMs (Setting 2). Generators are 
excluded from training.}
\label{tab:unseen}
\end{table*}

%% file: table/result3.tex
\begin{table*}[h]
\centering

\begin{tabular}{llccc}
\toprule
\textbf{Detector} & \textbf{Perturbation} 
  & \textbf{Accuracy (\%)} 
  & \textbf{FPR (\%)} 
  & \textbf{FNR (\%)} \\
\midrule

\multirow{3}{*}{LASTDE}
  & Paraphrasing   &49.45	&19.61	&53.77     \\
  & Humanization          & 54.16&	33.33	&47.14     \\
  & Back-translation & 66.73	&27.45	&33.88    \\
\midrule

\multirow{3}{*}{Likelihood}
  & Paraphrasing &  67.53&	27.45	&32.99      \\
  & Humanization       &  66.36	&19.61	&35.10     \\
  & Back-translation &74.31	&25.49	&25.71   \\
\midrule

\multirow{3}{*}{DetectAnyLLM}
  & Paraphrasing    &61.76	&56.86	&19.61      \\
  & Humanization      & 61.76	&56.86	&19.61    \\
  & Back-translation &63.73	&50.98	&21.57    \\
\midrule

\multirow{3}{*}{DeTeCtive}
  & Paraphrasing   &80.63	&45.10	&16.70    \\
  & Humanization      &  66.73	&49.02	&31.63     \\
  & Back-translation &78.74	&37.25	&19.59    \\
\midrule

\multirow{3}{*}{SeqXGPT}
  & Paraphrasing   &70.48	&9.80	&31.57   \\
  & Humanization     &   71.90	&29.41	&27.96      \\
  & Back-translation   &87.43	&19.61	&11.84     \\
\midrule

\multirow{3}{*}{CoCo}
  & Paraphrasing      & 91.55 & 84.75 &  1.67 \\
  & Humanization      & \textbf{89.22} & 81.54 &  4.76 \\
  & Back-translation  & 89.55 & 76.17 &  3.52 \\
\midrule
\multirow{3}{*}{\textbf{CSFG (Ours)}}
  & Paraphrasing      & \textbf{90.59} & 88.24          &  1.22 \\
  & Humanization      & 87.43          & 84.31          &  5.10          \\
  & Back-translation  & \textbf{90.96} &49.80 &  4.79          \\
\bottomrule
\end{tabular}
\caption{Detection performance under text perturbation (Setting 3) on Claude Sonnet 4.6. Results for GPT-5 and Gemini 2.5 Flash Lite are provided in Appendix~\ref{app:full}.}
\label{tab:robustness}
\end{table*}

%% file: section/conclusion.tex
\section{Conclusion}
We propose CSFG (Cross-Source Stylometric Fingerprint Graph), a graph-based AIGT detection framework grounded in \textit{Relational Over-Regularization} (ROR), the observation that LLMs produce recurring inter-sentence similarity
bursts that inflate and pattern-skew transition variance in a manner systematically distinguishable from human writing. We operationalize this as a learnable edge feature $\delta_{ij}$ within an edge-featured GNN, enabling end-to-end detection without manual thresholds.

Across three evaluation settings, CSFG achieves 97.14\% binary detection accuracy (+11.14pp over CoCo, FPR\,=\,1.57\%), generalizes to unseen LLMs with near-zero FPR, and reveals that LLM-based perturbation contaminates human text with generator-specific transition structure, an effect attributable to the rewriting process rather than detector failure. Ablation studies confirm that $\delta_{ij}$ acts as a false-positive calibrator rather than a raw detection booster.

CSFG fails structurally on GPT-5 (FNR\,=\,61.64\%), whose transition variance falls below the human baseline, a pattern we term hyper-uniform generation. This failure is not addressable by threshold adjustment; it reflects a fundamental boundary of ROR-based detection, which targets the presence of similarity bursts rather than the absence of organic irregularity, and will affect any detector trained exclusively on inflated-variance generators.

Future work may incorporate complementary discourse signals (e.g., rhetorical relations, argument structure) to extend coverage to hyper-uniform generators, explore sparse or hierarchical graph construction for scalability, and develop perturbation-aware training protocols to mitigate the contamination effect identified in Setting~3.

%% file: section/limitation.tex
\section*{Limitations}
\label{sec:limitations}

\paragraph{Scope of ROR and robustness to future generators.}
The ROR hypothesis may not hold consistently across all generators and writing domains. As demonstrated by the GPT-5 results in Setting~2 (FNR\,=\,61.64\%), certain LLMs produce transition-variance patterns that closely resemble or fall below the human baseline, rendering the similarity-burst signature an unreliable discriminative signal for those generators.
This boundary is structural: because $\delta_{ij}$ primarily targets the \emph{presence} and distribution of similarity bursts characteristic of the inflated-variance regime, it may not reliably distinguish generators whose relational patterns deviate in the opposite direction. Any detector trained exclusively on inflated-variance generators will face the same constraint. Addressing this limitation may require \emph{bidirectional relational modeling} that explicitly captures both inflated and suppressed transition patterns, together with complementary discourse-level signals beyond $\delta_{ij}$.
More broadly, the hypothesis may not generalize to creative writing or high-temperature generation, where idiosyncratic bursts differ in kind from the templated transitions our model targets, nor to highly structured human-written documents such as academic papers or news articles, which naturally exhibit periodic spikes in similarity---a prediction confirmed empirically in Appendix~\ref{app:full}, where FPR on news text (XSum) reaches 25\%, and where short documents (under 500 characters) similarly destabilize $\delta_{ij}$ estimation, raising FPR to 64--71\% for two of three unseen generators.

\paragraph{Perturbation-induced contamination.}
Robustness evaluation (Setting~3) reveals that LLM-based perturbation introduces a systematic side effect: rewriting human text with the same model used to generate AIGT imposes generator-specific transition patterns, inflating FPR attributable to rewriting rather than detector failure. Critically, this contamination effect is not specific to CSFG---CoCo exhibits similarly elevated FPR under the same conditions---suggesting it reflects a property of LLM rewriting rather than a detector deficiency. Future robustness evaluations should control for this effect by using a rewriting model distinct from the generator, and perturbation-aware training protocols may be necessary to mitigate it.

\paragraph{Sensitivity to representations and computational scalability.}
The framework relies on sentence-level cosine similarity computed over \texttt{roberta-base} CLS embeddings, which capture only part of the discourse structure and do not represent higher-level properties such as rhetorical organization or argument flow. Furthermore, pretrained encoders may share representational biases with AIGT, potentially attenuating the discriminative signal of $\delta_{ij}$ for certain generators. Regarding scalability, semantic edge computation scales quadratically with document length; future work may explore sparse connectivity or hierarchical graph construction to
address this constraint without sacrificing the relational signal captured by $\delta_{ij}$.

\section*{The Use of Large Language Models}
We used AI-assisted tools during the writing process for this manuscript. Specifically, we employed Grammarly for grammar checking and Claude-sonnet 4.6 for language polishing and to improve clarity of expression. These tools were used for editorial purposes.

%% file: section/appendix.tex
\newpage
\appendix
 
\section{Environment}
\label{sec:impl}
\paragraph{Hardware configuration:} The experiments were conducted on a system with an AMD Ryzen Threadripper 3960X 24-Core Processor and four NVIDIA RTX A6000 GPUs. The four NVIDIA RTX A6000 GPUs are used to train existing detectors.
 
\paragraph{LLM APIs:}  All text generation and perturbation experiments are conducted via the \textbf{OpenRouter API}. For Setting~2, we access \textbf{Claude Sonnet 4.6} (Anthropic), \textbf{GPT-5} (OpenAI), and \textbf{Gemini 2.5 Flash Lite} (Google). For Setting~3, perturbation prompts are applied to \textbf{Claude Sonnet 4.6}-generated text. All models are queried with a fixed random seed and temperature $= 1.0$ to ensure reproducibility.

\paragraph{Implementation Details}

Sentence embeddings are produced by \texttt{roberta-base}, encoding each sentence via the \texttt{[CLS]} token with 
$\ell_2$ normalisation (hidden dimension $768$, \texttt{max\_length}$=128$, batch size $128$). The encoder weights are frozen throughout training. Graphs are pre-built and cached to disk prior to training.

The GNN comprises $L = 3$ EdgeConv layers with hidden dimension $d_h = 128$ and dropout rate $0.2$. The model is optimised with AdamW (learning rate $3 \times 10^{-4}$, weight decay $5 \times 10^{-4}$) with batch size $32$. All experiments use a 
$70/15/15$ stratified train/validation/test split, and the checkpoint with the highest validation AUROC is selected for 
test evaluation.

Following \citet{common}, we use up to 2,000 samples per generator for Setting~1; Settings~2 and~3 use 1,000 samples in total.

\section{Ablation Study}
\label{app:ab}

\subsection{Effect of Auxiliary Loss Weight ($\lambda$)}
\label{app:ab1}
\input{table/ab1}
The joint training objective combines the binary detection loss $\mathcal{L}_\text{bin}$ and the source attribution loss $\mathcal{L}_\text{src}$ via a scalar weight $\lambda$. The source attribution head is designed to prevent the shared encoder from collapsing to shallow, generator-specific surface cues; however, an excessively large $\lambda$ risks over-regularising the encoder toward source discrimination at the expense of binary detection. To quantify this trade-off, we train CSFG with $\lambda \in \{0.0, 0.1, 0.3, 0.5, 1.0\}$, where $\lambda = 0.0$ corresponds to training without the auxiliary objective. All other hyperparameters are held at their default values.

Table~\ref{tab:lambda} reports the results. Removing the source attribution objective entirely ($\lambda = 0.0$) yields the highest FPR (2.27\%) and FNR (5.23\%), confirming that the auxiliary task provides meaningful regularisation for both error types. As $\lambda$ increases from 0.0 to 1.0, accuracy improves monotonically from 96.25\% to 97.22\%, and both FPR and FNR decline consistently, indicating that stronger source attribution pressure continues to benefit binary detection throughout the evaluated range. Unlike configurations where an excessively large $\lambda$ competes with the primary detection objective, performance does not degrade at $\lambda = 1.0$, suggesting that the two objectives are complementary rather than competing at this scale. We therefore adopt $\lambda = 1.0$ as the default configuration.

\subsection{Sensitivity to Semantic Edge Threshold ($\theta$)}
\label{app:ab2}
 
\input{table/ab2}
The semantic edge threshold $\theta$ controls the sparsity of long-range edges in the document graph: a low $\theta$ yields a dense graph in which the sequential backbone is supplemented by many weak semantic links, whereas a high $\theta$ produces a near-sequential graph that captures only strongly similar sentence pairs. We vary $\theta \in \{0.3, 0.4, 0.5, 0.6, 0.7\}$ and report detection performance across all three metrics.

Table~\ref{tab:theta} reports the results. Performance is notably stable across the evaluated range: accuracy varies by only 0.09pp (97.14\%-97.23\%) and FNR by 0.13pp (4.01\%-4.16\%), indicating that CSFG is not sensitive to the precise choice of $\theta$. The lowest FPR is observed at $\theta = 0.7$ (1.45\%), while the highest accuracy and lowest FNR are achieved at $\theta = 0.6$ (97.23\%, 4.01\%). FPR exhibits a non-monotonic pattern across thresholds, with no single value dominating all three metrics simultaneously.

We adopt $\theta = 0.6$ as the default configuration on the basis of validation-split AUROC, which shows a consistent advantage for this value across training runs. At $\theta = 0.6$, the graph connects sentence pairs with meaningfully high semantic similarity while avoiding the spurious long-range edges introduced by lower thresholds, aligning with the theoretical motivation of capturing genuine thematic recurrence rather than noisy co-occurrence. The marginal FPR advantage of $\theta = 0.7$ ($\Delta$FPR = 0.08pp) does not outweigh the accuracy and FNR gains at the default setting.

\subsection{Contribution of Edge Feature Components}
\label{app:ab3}
\input{table/ab3}

The four-dimensional edge feature vector $\mathbf{e}_{ij} = [\,\mathrm{sim}(s_i, s_j),\; \tilde{d}_{ij},\; \mathbb{1}_\mathrm{seq},\; \delta_{ij}\,]$ encodes complementary relational signals: semantic proximity, positional distance, edge type, and transition deviation. To assess the individual contribution of each component, we ablate them independently by zeroing out the corresponding index in $\mathbf{e}_{ij}$ at collation time, keeping the model architecture and graph structure unchanged.

Table~\ref{tab:edge} reports the effect of removing each edge feature component individually. The full model achieves the highest accuracy (97.14\%) across all variants, confirming that all four components contribute complementary signal. 
Among the ablated configurations, removing positional distance ($\tilde{d}_{ij}$) produces the largest degradation: accuracy drops to 96.89\% ($\Delta = -0.25$pp) and FNR rises to 4.67\%, the highest among all variants, indicating that position-aware relational encoding is the most critical component for distinguishing discourse topology between human and AIGT. Removing the sequential edge indicator ($\mathbb{1}_\mathrm{seq}$) and cosine similarity similarly degrade accuracy by 0.20pp and 0.22pp respectively, and raise FNR to 4.42\% and 4.57\%, confirming that each component carries non-redundant discriminative signal.

The result for \textit{w/o trans\_dev} warrants careful interpretation. Removing $\delta_{ij}$ yields a higher accuracy than the \textit{w/o sim} and \textit{w/o dist} variants (96.97\%) and the lowest FNR among all ablations (4.40\%), yet 
its FPR (1.65\%) exceeds that of the full model (1.57\%). This pattern confirms the expected calibrating role of $\delta_{ij}$: rather than maximising raw detection rate, it acts as a regulariser on false positives, suppressing the 
tendency to classify borderline human-written text as AI-generated. The full model achieves the most balanced error profile across all three metrics, which is preferable in deployment settings where falsely accusing human authors carries 
a higher social cost than missing AIGT.

\subsection{Effect of GNN Depth}
\label{app:ab4}
\input{table/ab4}

The number of EdgeConv layers determines the receptive field of each node in the sentence graph: a $k$-layer network aggregates relational signals from sentences up to $k$ hops away. In a sentence-level graph, one hop corresponds to an adjacent or semantically linked sentence, so three layers approximate the span of a coherent paragraph-level discourse unit. Deeper networks risk over-smoothing, in which node representations converge and lose local distinctiveness, while shallower networks may fail to capture document-level topology. We compare configurations with 1, 2, 3, and 4 EdgeConv layers, keeping all other settings fixed at their default values ($\theta = 0.6$, $\lambda = 1.0$).

Table~\ref{tab:depth} reports detection performance as a function of GNN depth. The single-layer network yields the lowest accuracy (96.59\%) and the highest FPR (2.06\%) and FNR (4.75\%), indicating that a receptive field limited to one sentence hop is insufficient to capture the document-level discourse topology required for reliable detection. 
Performance improves substantially at two layers (97.09\%), with both FPR (1.53\%) and FNR (4.29\%) declining, confirming that aggregating relational context beyond immediate neighbours provides meaningful discriminative signal.

Three layers yield the best accuracy (97.15\%) and the lowest FPR (1.42\%), supporting the design rationale that aggregating relational context up to three sentence hops approximates a coherent paragraph-level discourse unit. 
Extending to four layers produces marginal performance degradation in accuracy (97.10\%) and a slight FPR increase (1.53\%), consistent with the over-smoothing phenomenon in which deeper message passing causes node representations to converge and lose local discriminative structure. FNR continues to decrease marginally beyond three layers (4.27\% $\to$ 4.26\%), but the gain of 0.01pp does not justify the accompanying increase in false positives. Three layers therefore strikes the best balance across all three metrics and is adopted as the default configuration.

\section{Generator-Level Transition Variance Analysis}
\label{app:generator_tv}

Section~5.3 reports that CSFG achieves a false negative rate of 61.64\% on GPT-5 while attaining FNR $\leq$ 20.46\% on all other unseen generators. To determine whether this failure reflects a detector deficiency or a qualitative difference in the generative behaviour of GPT-5, we perform a generator-level analysis of transition variance $\mathrm{Var}(T(D))$ on the
Setting~2 test set.

\subsection{Per-Generator Transition Variance Statistics}

Table~\ref{tab:tv_generator} reports $\mathrm{Var}(T(D))$ statistics for each group alongside the corresponding Setting~2 FNR. Statistical significance is assessed via the Mann-Whitney $U$ test against the human reference group; distributional distance is measured by the Kolmogorov-Smirnov (KS) statistic $D$.

\begin{table*}[h]
\centering

\begin{tabular}{lcccccc}
\toprule
\textbf{Group} & $n$ & \textbf{TV mean} & \textbf{TV std}
               & \textbf{$p$-value} & \textbf{KS $D$} & \textbf{FNR (\%)} \\
\midrule
Human            & 500 & $1.1\times10^{-3}$ & $6\times10^{-4}$ & --              & --    & -- \\
GPT-5            & 500 & $0.8\times10^{-3}$ & $5\times10^{-4}$ & $6.05\times10^{-13}$  & 0.2389 & 61.64 \\
Claude Sonnet 4.6 & 500 & $1.5\times10^{-3}$ & $5\times10^{-4}$ & $1.09\times10^{-19}$  & 0.2915 & 20.46 \\
Gemini 2.5 Flash Lite & 500 & $2.3\times10^{-3}$ & $6\times10^{-4}$ & $8.88\times10^{-102}$ & 0.6482 & 14.81 \\
\bottomrule
\end{tabular}
\caption{Per-generator transition variance statistics on the Setting~2 test set. TV mean and TV std denote the mean and standard deviation of $\mathrm{Var}(T(D))$ per group. $p$-value: Mann-Whitney $U$ test against the human group. KS $D$: Kolmogorov-Smirnov distance against the human CDF. FNR: miss rate of CSFG from Table~3.}
\label{tab:tv_generator}
\end{table*}

\subsection{GPT-5: Hyper-Uniform Generation}
\label{app:gpt5_train}
\begin{figure}[h]
\centering
\includegraphics[width=0.72\linewidth]{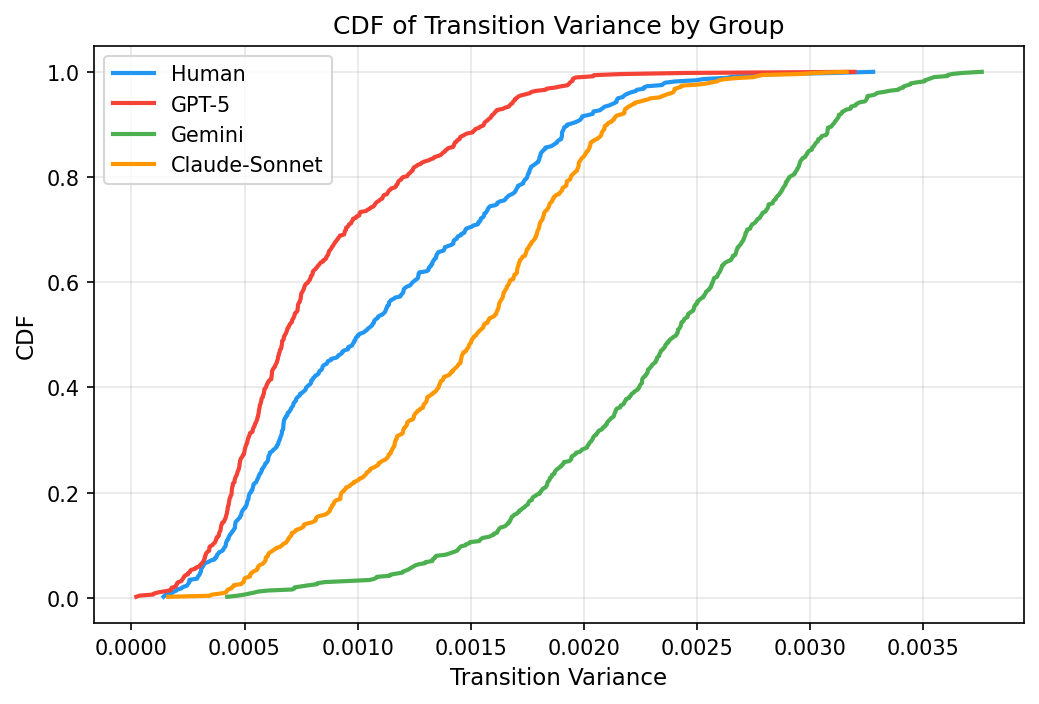}
\caption{Cumulative distribution functions of $\mathrm{Var}(T(D))$ for human-written text and three unseen generators (Setting~2 test set). GPT-5 (red) lies to the \emph{left} of human text (blue), indicating hyper-uniform generation with suppressed transition variance. Gemini 2.5 Flash Lite (green) is displaced far to the right, reflecting an amplified ROR signal. Claude Sonnet 4.6 (orange) occupies an intermediate position. The close overlap between GPT-5 and human CDFs (KS $D = 0.2389$) directly explains the elevated FNR reported in Table~3.}
\label{fig:tv_cdf}
\end{figure}

As shown in Table~\ref{tab:tv_generator} and Figure~\ref{fig:tv_cdf}, GPT-5 exhibits a mean transition variance of $0.8\times10^{-3}$, \emph{lower} than that of human-written text ($1.1\times10^{-3}$).
This reverses the directional prediction of the ROR hypothesis, under which AIGT is expected to exhibit \emph{higher} transition variance than human writing.
We term this pattern \textbf{hyper-uniform generation}: inter-sentence similarity is suppressed below the organic irregularity of human discourse, rather than being inflated by recurring similarity bursts.

The KS distance between GPT-5 and human transition variance distributions ($D = 0.2389$) is less than half that of Gemini 2.5 Flash Lite ($D = 0.6482$), and the CDF of GPT-5 closely overlaps with that of human text across the bulk of the distribution (Figure~\ref{fig:tv_cdf}). This near-indistinguishability places GPT-5 outside the high-variance decision region learned from training generators, directly explaining the elevated FNR of 61.64\%.

Critically, the failure is structural rather than a threshold calibration issue: because $\delta_{ij}$ is defined as the signed deviation from the document-level mean similarity, a hyper-uniform document produces uniformly small $|\delta_{ij}|$ values that do not trigger the burst signature targeted by the ROR hypothesis. Simple threshold adjustment cannot resolve this, as lowering the detection boundary would simultaneously increase false positives on human text.
Reliable detection of hyper-uniform generators likely requires either retraining with GPT-5 examples or incorporating complementary features that capture the \emph{absence} of organic irregularity rather than the \emph{presence} of similarity bursts.

\subsection{Gemini 2.5 Flash Lite: Amplified ROR Signal}

In contrast, Gemini 2.5 Flash Lite exhibits a mean transition variance of $2.3\times10^{-3}$, more than twice that of human text, with a KS distance of 0.6482 (Figure~\ref{fig:tv_cdf}, green curve). This amplified ROR signal—arising from markedly abrupt topic transitions between sentences—is strongly discriminable by CSFG, consistent with the low FNR of 14.81\% observed in Setting~2.

\subsection{Implications for ROR-Based Detection}

Taken together, these results reveal that the transition variance landscape of frontier LLMs is not unimodal.
Generators differ qualitatively in the direction of their deviation from human transition variance: some amplify the ROR burst pattern (Gemini), some partially exhibit it (Claude Sonnet 4.6), and some invert it (GPT-5).
The ROR hypothesis is most predictive for generators whose training or decoding procedure introduces recurring similarity spikes; it is least applicable to generators whose outputs fall within or below the human variance range.
Future work should investigate whether hyper-uniform generation is a deliberate property of GPT-5's training objective or an incidental effect of scale, and whether it will generalize to subsequent model generations.

\section{Full Results of Setting 3}
\label{app:full}
\input{table/ab_exp3}

\subsection{Document Length and Domain Structure}
\label{sec:boundary}

The evaluations in \S5.2--\S5.4 use documents with moderate-to-long sentence counts, providing sufficient sentence pairs for estimating $\delta_{ij}$ and $\mathrm{Var}(T(D))$. We examine two boundary conditions where this assumption may break down: (1) \emph{short documents}, where too few sentence pairs may yield unstable transition-variance estimates, and (2) \emph{highly structured human writing}, where conventional discourse patterns may produce similarity spikes resembling the ROR signature.

\paragraph{Short-text performance.}
We construct a subset of documents shorter than 500 characters from the Setting~2 generators and evaluate the Setting~1 model without retraining. Table~\ref{tab:short-text} summarizes the results.
 
\begin{table*}[t]
\centering

\begin{tabular}{lrrrr}
\toprule
\textbf{Generator} & \textbf{$N$} & \textbf{Acc.\ (\%)} & \textbf{FPR (\%)} & \textbf{FNR (\%)} \\
\midrule
Claude Sonnet 4.6 & 125 & 56.00 & 63.86 & 4.76 \\
Gemini 2.5 Flash Lite & 125 & 51.20 & 71.08 & 4.76 \\
GPT-5 & 125 & 74.40 & 18.07 & 40.48 \\
\bottomrule
\end{tabular}
\caption{Detection performance on documents shorter than 500 characters. FPR increases sharply relative to the 0.66\% observed on the full-length Setting~1 test set, while FNR remains low for Claude Sonnet~4.6 and Gemini~2.5~Flash~Lite but increases substantially for GPT-5.}
\label{tab:short-text}
\end{table*}
 
Short documents substantially degrade detection performance, particularly by increasing false positives. For Claude Sonnet~4.6 and Gemini~2.5~Flash~Lite, FPR rises by more than an order of magnitude relative to the full-length setting. This behavior follows directly from the construction of $\delta_{ij}$: with fewer sentences, the number of sentence pairs $\binom{n}{2}$ decreases rapidly, making the document-level mean similarity $\bar{\sigma}$ more sensitive to individual sentence pairs. Consequently, a single locally similar pair can disproportionately influence the estimated transition-variance pattern.

The resulting low FNR but high FPR suggests that the detector tends to classify uncertain short documents as AI-generated. This is consistent with a decision boundary calibrated on longer documents, where positive $\delta_{ij}$ values more reliably indicate genuine similarity bursts rather than sampling variation. GPT-5 exhibits a different failure mode. Because its transition variance is already low (Appendix~\ref{app:full}), short-document sampling does not produce the same inflated-variance pattern. Its FPR therefore remains comparatively lower, but FNR increases to 40.48\%, indicating reduced sensitivity to GPT-5's hyper-uniform transition structure.

\paragraph{Structured human domains.}
We further evaluate CSFG on 100 human-written documents from XSum \citep{xsum}, a collection of professionally edited BBC news articles characterized by a relatively formulaic discourse structure, including lead sentences, inverted-pyramid organization, and recurring topic restatement. Table~\ref{tab:xsum} reports the results.
 
\begin{table}[!h]
\centering
\small
\begin{tabular}{lrrrr}
\toprule
\textbf{Dataset} & \textbf{$N$} & \textbf{Acc.\ (\%)} & \textbf{FPR (\%)} & \textbf{FNR (\%)} \\
\midrule
XSum & 100 & 85.00 & 25.00 & 6.00 \\
\bottomrule
\end{tabular}
\caption{CSFG performance on human-written news text (XSum). FPR is substantially elevated relative to the 1.57\% observed on the balanced Setting~1 test set.}
\label{tab:xsum}
\end{table}
 
CSFG produces an FPR of 25.00\% on XSum, substantially higher than the 1.57\% observed on the balanced Setting~1 test set. This result demonstrates a domain-specific confound: professionally edited news articles naturally contain paragraph-boundary restatements and recurring transition patterns that can resemble the similarity bursts targeted by ROR. Thus, the relational signature is not inherently unique to AIGT; it can also arise from conventionalized human discourse structures.

This failure mode cannot be addressed simply by adding more LLM generators to the training set, because the error stems from the interaction between ROR and a particular human writing convention rather than generator-specific variation. Mitigating it will likely require domain-aware calibration or complementary discourse features that distinguish AI-generated regularities from legitimate structural regularities in human writing.

Overall, these results identify two important boundary conditions for CSFG. Short documents provide too few relational observations for stable estimation, whereas highly structured human genres can produce relational patterns that resemble ROR. Accordingly, CSFG's low FPR should not be assumed to generalize uniformly across document lengths or writing domains. We report these cases explicitly to characterize when the proposed relational signal becomes unreliable.

\subsection{Cross-Generator Comparison}
\label{sec:appendix-cross-generator}

\begin{table*}[t]
\centering
\small
\setlength{\tabcolsep}{4pt}
\begin{tabular}{ll ccc ccc ccc}
\toprule
& & \multicolumn{3}{c}{SeqXGPT} & \multicolumn{3}{c}{CoCo} & \multicolumn{3}{c}{CSFG (Ours)} \\
\cmidrule(lr){3-5} \cmidrule(lr){6-8} \cmidrule(lr){9-11}
Generator & Perturbation & Acc & FPR & FNR & Acc & FPR & FNR & Acc & FPR & FNR \\
\midrule
\multirow{3}{*}{Claude Sonnet 4.6}
 & Paraphrasing    & 70.48 & 9.80  & 31.57 & 91.55 & 84.75 & 1.67 & 90.59 & 88.24 & 1.22 \\
 & Humanization     & 71.90 & 29.41 & 27.96 & 89.22 & 81.54 & 4.76 & 87.43 & 84.31 & 5.10 \\
 & Back-translation & 87.43 & 19.61 & 11.84 & 89.55 & 76.17 & 3.52 & 90.96 & 49.80 & 4.79 \\
\midrule
\multirow{3}{*}{Gemini 2.5 Flash Lite}
 & Paraphrasing    & 58.30 & 7.84  & 45.21 & 89.92 & 69.51 & 3.53 & 85.98 & 92.16 & 5.91 \\
 & Humanization     & 64.64 & 37.25 & 35.16 & 79.35 & 76.17 & 14.83 & 76.43 & 82.35 & 17.48 \\
 & Back-translation & 89.50 & 9.80  & 10.57 & 90.91 & 95.92 & 1.24 & 91.53 & 37.25 & 5.49 \\
\midrule
\multirow{3}{*}{GPT-5}
 & Paraphrasing    & 80.53 & 14.29 & 20.00 & 84.04 & 60.25 & 11.92 & 77.86 & 36.73 & 20.63 \\
 & Humanization     & 71.07 & 20.00 & 29.84 & 59.05 & 34.44 & 41.62 & 47.93 & 20.00 & 55.35 \\
 & Back-translation & 77.09 & 6.45  & 24.46 & 88.37 & 66.54 & 6.64 & 86.31 & 22.58 & 12.84 \\
\bottomrule
\end{tabular}
\caption{Cross-generator summary of Setting 3 robustness results (\%). FPR is measured on perturbed human-written text. Detailed per-generator results are reported separately in Table~\ref{tab:robustness}, \ref{tab:full1}, and \ref{tab:full2}.}
\label{tab:setting3-cross-generator-summary}
\end{table*}

Table~\ref{tab:setting3-cross-generator-summary} summarizes the Setting~3 results across all three unseen generators, drawing together the per-generator results reported in Table~\ref{tab:robustness}, \ref{tab:full1}, and \ref{tab:full2}. Two patterns generalize across generators. First, CSFG and CoCo both exhibit substantially elevated FPR under paraphrasing and humanization relative to back-translation, consistent with the contamination effect discussed in \S5.4: LLM-based rewriting imposes generator-specific transition structure onto human text regardless of the source generator. Second, back-translation consistently yields the most favorable FPR/FNR trade-off for CSFG across all three generators (49.80\%, 37.25\%, and 22.58\% FPR for Claude, Gemini, and GPT-5, respectively), supporting our claim that it better preserves the original transition structure than direct stylistic rewriting.

GPT-5 humanization is a notable exception to this trend. Under this condition, CSFG's accuracy drops to 47.93\%, with FNR rising to 55.35\%—below SeqXGPT's accuracy (71.07\%) on the same condition and substantially worse than CSFG's own performance on Claude and Gemini humanization. This compounds the hyper-uniform generation behavior of GPT-5 identified in Appendix~C: because GPT-5's unperturbed transition variance already falls near or below the human baseline, humanization-induced rewriting appears to push a large fraction of GPT-5-generated documents further into the region CSFG associates with human text, rather than introducing the burst-like artifacts that paraphrasing or back-translation partially retain. This result reinforces the boundary condition noted in \S6 and Appendix~\ref{app:generator_tv}: ROR-based detection is structurally weakest for generators whose transition-variance regime departs from the inflated-variance direction, and this weakness is amplified, not mitigated, by humanization-style perturbation.

\section{LLM Prompt}

\label{app:prompt}

All text generation and perturbation experiments use the OpenRouter API. Table~\ref{tab:prompts} lists the prompts used for each condition.

For Setting~2 (unseen model generalization), we adopt the generation prompt from \citet{detectgpt}, which instructs the model to produce a coherent continuation given the first 30 tokens of a document.

For Setting~3 (robustness evaluation), we employ three perturbation prompts drawn from prior work: the paraphrase prompt follows \citet{paraphrase}, the humanization prompt follows \citet{human}, and the back-translation prompt follows \citet{back}, which performs a two-stage English$\to$Chinese$\to$English pipeline to introduce surface-level variation while preserving semantic content.
  
\input{table/prompt}

%% file: table/ab1.tex
\begin{table*}[h]
\centering

\begin{tabular}{lcccccc}
\toprule
\multirow{2}{*}{$\lambda$} 
  & \multicolumn{2}{c}{\textbf{Accuracy (\%)}} 
  & \multicolumn{2}{c}{\textbf{FPR (\%)}} 
  & \multicolumn{2}{c}{\textbf{FNR (\%)}} \\
\cmidrule(lr){2-3}\cmidrule(lr){4-5}\cmidrule(lr){6-7}
  & $\mu$ & $\sigma$ & $\mu$ & $\sigma$ & $\mu$ & $\sigma$ \\
\midrule
0.0 (no aux.) & 96.25 & 0.42 & 2.27 & 0.31 & 5.23 & 0.81 \\
0.1           & 96.95 & 0.26 & 1.68 & 0.53 & 4.42 & 0.83 \\
0.3           & 97.02 & 0.24 & 1.51 & 0.40 & 4.44 & 0.58 \\
0.5           & 97.19 & 0.17 & 1.60 & 0.38 & 4.02 & 0.46 \\
\textbf{1.0} & \textbf{97.22} & 0.22 & \textbf{1.48} & 0.39 & \textbf{4.08} & 0.45 \\
\bottomrule
\end{tabular}
\caption{Effect of auxiliary loss weight $\lambda$ on binary detection 
performance. All results are averaged over 10 runs; $\sigma$ denotes 
standard deviation in percentage points (pp). \emph{No aux.}\ ($\lambda = 
0.0$) denotes training without the source attribution head. Bold indicates 
the best value per column.}
\label{tab:lambda}
\end{table*}

%% file: table/ab2.tex
\begin{table*}[h]
\centering

\begin{tabular}{lcccccc}
\toprule
\multirow{2}{*}{$\theta$} 
  & \multicolumn{2}{c}{\textbf{Accuracy (\%)}} 
  & \multicolumn{2}{c}{\textbf{FPR (\%)}} 
  & \multicolumn{2}{c}{\textbf{FNR (\%)}} \\
\cmidrule(lr){2-3}\cmidrule(lr){4-5}\cmidrule(lr){6-7}
  & $\mu$ & $\sigma$ & $\mu$ & $\sigma$ & $\mu$ & $\sigma$ \\
\midrule
0.3           & 97.20 & 0.24 & 1.48 & 0.30 & 4.11 & 0.55 \\
0.4           & 97.22 & 0.21 & 1.51 & 0.29 & 4.05 & 0.51 \\
0.5           & 97.14 & 0.21 & 1.58 & 0.36 & 4.14 & 0.48 \\
\textbf{0.6} & \textbf{97.23} & 0.17 & 1.53 & 0.43 & \textbf{4.01} & 0.45 \\
0.7           & 97.19 & 0.22 & \textbf{1.45} & 0.30 & 4.16 & 0.44 \\
\bottomrule
\end{tabular}
\caption{Sensitivity to semantic edge threshold $\theta$. Lower $\theta$ 
yields denser graphs with more long-range semantic edges. All results are 
averaged over 10 runs; $\sigma$ denotes standard deviation in percentage 
points (pp). Bold indicates the best value per column.}
\label{tab:theta}
\end{table*}

%% file: table/ab3.tex
\begin{table*}[t]
\centering

\begin{tabular}{lcccccc}
\toprule
\multirow{2}{*}{\textbf{Edge Features}} 
  & \multicolumn{2}{c}{\textbf{Accuracy (\%)}} 
  & \multicolumn{2}{c}{\textbf{FPR (\%)}} 
  & \multicolumn{2}{c}{\textbf{FNR (\%)}} \\
\cmidrule(lr){2-3}\cmidrule(lr){4-5}\cmidrule(lr){6-7}
  & $\mu$ & $\sigma$ & $\mu$ & $\sigma$ & $\mu$ & $\sigma$ \\
\midrule
\textbf{Full (ours)}  & \textbf{97.14} & 0.27 & \textbf{1.57}  & 0.38 & \textbf{4.15} & 0.54 \\
w/o sim               & 96.92 & 0.21 & 1.59 & 0.49 & 4.57 & 0.66 \\
w/o dist              & 96.89 & 0.24 & 1.54 & 0.51 & 4.67 & 0.76 \\
w/o is\_seq           & 96.94 & 0.17 & 1.69 & 0.45 & 4.42 & 0.61 \\
w/o trans\_dev        & 96.97 & 0.23 & 1.65 & 0.34 & 4.40 & 0.67 \\
\bottomrule
\end{tabular}
\caption{Contribution of each edge feature component. Each variant zeros 
out the corresponding dimension of $\mathbf{e}_{ij}$ at collation time; 
graph structure and model architecture remain unchanged. All results are 
averaged over 10 runs; $\sigma$ denotes standard deviation in percentage 
points (pp). Bold indicates the best value per column.}
\label{tab:edge}
\end{table*}

%% file: table/ab4.tex
\begin{table*}[t]
\centering

\begin{tabular}{lcccccc}
\toprule
\multirow{2}{*}{\textbf{\# Layers}} 
  & \multicolumn{2}{c}{\textbf{Accuracy (\%)}} 
  & \multicolumn{2}{c}{\textbf{FPR (\%)}} 
  & \multicolumn{2}{c}{\textbf{FNR (\%)}} \\
\cmidrule(lr){2-3}\cmidrule(lr){4-5}\cmidrule(lr){6-7}
  & $\mu$ & $\sigma$ & $\mu$ & $\sigma$ & $\mu$ & $\sigma$ \\
\midrule
1                 & 96.59 & 0.19 & 2.06 & 0.31 & 4.75 & 0.63 \\
2                 & 97.09 & 0.26 & 1.53 & 0.28 & 4.29 & 0.71 \\
\textbf{3 (ours)} & \textbf{97.15} & 0.25 & \textbf{1.42} & 0.43 & 4.27 & 0.67 \\
4                 & 97.10 & 0.31 & 1.53 & 0.46 & \textbf{4.26} & \textbf{0.68} \\
\bottomrule
\end{tabular}
\caption{Effect of GNN depth on binary detection performance. All results 
are averaged over 10 runs; $\sigma$ denotes standard deviation in percentage 
points (pp). All variants use the full edge feature vector and default 
configuration ($\theta = 0.6$, $\lambda = 1.0$). Bold indicates the best 
value per column.}
\label{tab:depth}
\end{table*}

%% file: table/ab_exp3.tex
\begin{table*}[h]
\centering

\begin{tabular}{llccc}
\toprule
\textbf{Detector} & \textbf{Perturbation} 
  & \textbf{Accuracy (\%)} 
  & \textbf{FPR (\%)} 
  & \textbf{FNR (\%)} \\
\midrule

\multirow{3}{*}{LASTDE}
  & Paraphrasing      &  42.62	&9.80	&62.32  \\
  & Humanization      &   77.72	&56.86	&18.70        \\
  & Back-translation  &  54.14	&23.53	&48.17     \\
\midrule

\multirow{3}{*}{Likelihood}
  & Paraphrasing      &   60.70	&33.33	&39.92   \\
  & Humanization      &   46.41	&21.57	&56.91     \\
  & Back-translation  & 76.61	&27.45	&22.97    \\
\midrule

\multirow{3}{*}{DetectAnyLLM}
  & Paraphrasing      & 51.96	&78.43	&17.65    \\
  & Humanization      &    53.92	&49.02	&43.14         \\
  & Back-translation  &  66.67	&47.06	&19.61    \\
\midrule

\multirow{3}{*}{DeTeCtive}
  & Paraphrasing      &36.72	&5.88	&69.25  \\
  & Humanization      &  76.61	&66.67&	18.90       \\
  & Back-translation  &  79.37	&31.37	&19.51        \\
\midrule

\multirow{3}{*}{SeqXGPT}
  & Paraphrasing      &  58.30	&7.84	&45.21  \\
  & Humanization      & 64.64	&37.25	&35.16      \\
  & Back-translation  & 89.50	&9.80	&10.57        \\
\midrule

\multirow{3}{*}{CoCo}
  & Paraphrasing      &   0.90	&0.70	&0.04	   \\
  & Humanization      & 0.79	&0.76	&0.15        \\
  & Back-translation  &  0.91	&0.96	&0.01        \\
\midrule
\multirow{3}{*}{\textbf{CSFG (Ours)}}
  & Paraphrasing      &    0.86	&0.92	&0.06   \\
  & Humanization      &    0.76	&0.82	&0.17      \\
  & Back-translation  &  0.92	&0.37	&0.05           \\
\bottomrule
\end{tabular}
\caption{Detection performance under text perturbation (Setting 3) on Gemini 2.5 Flash Lite.}
\label{tab:full1}
\end{table*}


\begin{table*}[h]
\centering

\begin{tabular}{llccc}
\toprule
\textbf{Detector} & \textbf{Perturbation} 
  & \textbf{Accuracy (\%)} 
  & \textbf{FPR (\%)} 
  & \textbf{FNR (\%)} \\
\midrule

\multirow{3}{*}{LASTDE}
  & Paraphrasing      &     43.51	&22.45	&60.00    \\
  & Humanization      &  75.62	&60.00	&20.73       \\
  & Back-translation  &  70.11	&35.48	&29.36     \\
\midrule

\multirow{3}{*}{Likelihood}
  & Paraphrasing      &   65.84	&26.53	&34.95   \\
  & Humanization      &   38.02	&15.56	&66.74        \\
  & Back-translation  &   65.92	&16.13	&35.78     \\
\midrule

\multirow{3}{*}{DetectAnyLLM}
  & Paraphrasing      &  64.29	&51.02	&20.41   \\
  & Humanization      &  52.22	&26.67	&68.89         \\
  & Back-translation  &  61.29	&6.45	&70.97     \\
\midrule

\multirow{3}{*}{DeTeCtive}
  & Paraphrasing      &  72.52	&44.90	&25.68  \\
  & Humanization      &  52.27	&24.44&	50.11       \\
  & Back-translation  & 76.26	&54.84	&20.80    \\
\midrule

\multirow{3}{*}{SeqXGPT}
  & Paraphrasing      &  80.53	&14.29	&20.00    \\
  & Humanization      &   71.07	&20.00	&29.84      \\
  & Back-translation  &  77.09	&6.45	&24.46         \\
\midrule

\multirow{3}{*}{CoCo}
  & Paraphrasing      &   0.84	&0.60&	0.12   \\
  & Humanization      &     0.59	&0.34	&0.42           \\
  & Back-translation  &  0.88	&0.67	&0.07       \\
\midrule
\multirow{3}{*}{\textbf{CSFG (Ours)}}
  & Paraphrasing      &  0.78	&0.23	&0.13   \\
  & Humanization      &   0.48	&0.20	&0.55         \\
  & Back-translation  &    0.86	&0.37	&0.21         \\
\bottomrule
\end{tabular}
\caption{Detection performance under text perturbation (Setting 3) on GPT-5.}
\label{tab:full2}
\end{table*}

%% file: table/prompt.tex
\begin{table*}[h]
  \centering
  \renewcommand{\arraystretch}{1.4}
  \setlength{\tabcolsep}{8pt}
  \begin{tabularx}{\textwidth}{@{} c l X @{}}
    \toprule
    \textbf{Set.} & \textbf{Condition} & \textbf{Prompt} \\
    \midrule

    2 & Generation &
      \textit{Please provide a continuation for the following content
      to make it coherent: \{first 30 tokens\}} \\

    \midrule

    \multirow{4}{*}{3}
      & Paraphrase &
        \textit{Paraphrase the following text: \{Text\}} \\[4pt]

      & Humanization &
        \textit{Paraphrase the given sentence. Only return the
        paraphrased sentence in your response. Make it seem like a
        human wrote the article and that it is from the YOUR SECTION
        section of YOUR PUBLICATION.
        Sentence to paraphrase: YOUR SENTENCE
        Your paraphrase of the sentence: \{Text\}} \\[4pt]

      & \multirow[t]{2}{*}{Back-translation}
        & \textit{You are a professional translator. Translate the
          following English text into Chinese. Rules: (1)~Preserve the
          original meaning and tone faithfully. (2)~Do not add
          explanations or commentary. (3)~Output only the translated
          Chinese text.
          Text: \{input\_text\}} \\[2pt]

      & &
        \textit{You are a professional translator. Translate the
          following Chinese text back into English. Rules:
          (1)~Translate naturally---do not attempt to recover the
          original wording. (2)~Preserve meaning and tone.
          (3)~Output only the translated English text.
          Text: \{chinese\_text\}} \\

    \bottomrule
  \end{tabularx}
  \caption{Prompts used for text generation (Setting~2)
           and perturbation (Setting~3).}
  \label{tab:prompts}
\end{table*}